\def\year{2020}\relax
\documentclass[letterpaper]{article} 
\usepackage{aaai20}  
\usepackage{times}  
\usepackage{helvet} 
\usepackage{courier}  
\usepackage[hyphens]{url}  
\usepackage{graphicx} 
\usepackage{graphicx}  
\usepackage{array,amsmath,amssymb,amsthm,booktabs}
\newtheorem{theorem}{Theorem}
\newtheorem{lemma}[theorem]{Lemma}
\title{Unsupervised Domain Adaptation via Discriminative Manifold Embedding and Alignment}
\author{You-Wei Luo \textsuperscript{\rm 1},
Chuan-Xian Ren \textsuperscript{\rm 1}\thanks{Corresponding Author.},
Pengfei Ge \textsuperscript{\rm 1},
Ke-Kun Huang\textsuperscript{\rm 2},
Yu-Feng Yu\textsuperscript{\rm 3} \\
\textsuperscript{\rm 1} School of Mathematics, Sun Yat-Sen University, China\\
\textsuperscript{\rm 2} School of Mathematics, JiaYing University, China \\
\textsuperscript{\rm 3} Department of Statistics and Institute of Intelligent Finance, Guangzhou University, China \\
\{luoyw28, gepengf\}@mail2.sysu.edu.cn~~~~rchuanx@mail.sysu.edu.cn \\
kkcocoon@163.com~~~~yuyufeng220@163.com
}
 
\begin{document}

\maketitle

\begin{abstract}
Unsupervised domain adaptation is effective in leveraging the rich information from the source domain to the unsupervised target domain. Though deep learning and adversarial strategy make an important breakthrough in the adaptability of features, there are two issues to be further explored. First, the hard-assigned pseudo labels on the target domain are risky to the intrinsic data structure. Second, the batch-wise training manner in deep learning limits the description of the global structure. In this paper, a Riemannian manifold learning framework is proposed to achieve transferability and discriminability consistently. As to the first problem, this method establishes a probabilistic discriminant criterion on the target domain via soft labels. Further, this criterion is extended to a global approximation scheme for the second issue; such approximation is also memory-saving. The manifold metric alignment is exploited to be compatible with the embedding space. A theoretical error bound is derived to facilitate the alignment. Extensive experiments have been conducted to investigate the proposal and results of the comparison study manifest the superiority of consistent manifold learning framework.
\end{abstract}

\section{Introduction}\label{sec1}
In machine learning, large-scale datasets with annotations play a crucial role during the learning process. Convolutional Neural Networks (CNNs) achieves a significant advance in various tasks via a huge number of well-labeled samples \cite{DNN}. Unfortunately, such data is actually prohibitive in many real-world scenarios. Applying the learned model in the new environment, i.e., the cross-domains scheme, will cause a significant degradation of recognition performance \cite{ren2018generalized,kim2019unsupervised}.

Unsupervised Domain Adaptation (UDA) is designed to deal with the shortage of labels by leveraging the rich labels and strong supervision from the source domain to the target domain, where the target domain has no access to the annotations. In fact, datasets composed of specifically exploratory factors and variants, such as background, style, illumination, camera views or resolution, often lead to the shifting distributions (i.e., the domain shift) \cite{shimodaira2000improving,moreno2012unifying}. According to the transfer theory established by Ben-David et al. \cite{ben2007analysis,ben2010theory}, the primary task for cross-domain adaptation is to learn the discriminative feature representations while narrowing the discrepancy between domains.

Recent literature indicates that CNNs learn abstract representations with nonlinear transformations \cite{bengio2013representation}, which suppress the negative effects caused by variant explanatory factors in domain shift \cite{long2015learning}. Pioneer works \cite{long2015learning,ganin2016domain,Long2017Deep,sankaranarayanan2018generate} attempt to transfer the source classifier with sufficient supervision to the target domain by minimizing the discrepancy between the source and target domains. Though early adversarial confusion methods \cite{ganin2016domain,sankaranarayanan2018generate,pinheiro2018unsupervised}, which is inspired by Generative Adversarial Nets (GANs) \cite{goodfellow2014generative}, promise the generated features are domain-indistinguishable and form a well-aligned the marginal distributions, the conditional distributions are still not guaranteed \cite{long2018conditional,saito2018maximum,chen2019transferability}.

Some latest methods achieve remarkable improvement in accuracy by employing the uncertainty information on the target domain, e.g., pseudo labels and soft labels \cite{long2018conditional,saito2018maximum,pinheiro2018unsupervised,chen2019transferability}. Though such information transduced from the source domain strengthens the discriminative ability of the target domain, there are still two points to be further explored. First, direct utilization of uncertainty information is risky and should be treated cautiously \cite{long2018conditional}, as the hard-assigned pseudo labels may change the intrinsic structure of data space \cite{ding2019deep}. Second, the batch-wise training in deep learning limits the capture of global information; thus models may be misled by some extreme local distributions.

In this paper, we develop a novel Riemannian manifold embedding and alignment framework. As the transferability and discriminability are both valuable \cite{chen2019transferability}, the proposal reaches a consistent rule for these two properties. The main idea is to describe the domains by a sequence of abstract manifolds. Enlightened by the successful application of soft labels for conditional coding and the multilayer embedding in \cite{Long2017Deep,long2018conditional}, a probabilistic discriminant criterion is proposed. Further, we extend this criterion to a global approximation scheme, which overcomes the dilemma of discriminant learning in batch-wise training. Inspired by previous attempts on manifold learning \cite{gong2012geodesic,huang2017cross}, we employ manifold metric to measure the domain discrepancy. The contributions are summarized as follows.
\begin{itemize}
\item To optimize the structure of the target domain and reduce the risk of uncertainty information simultaneously, a probabilistic discriminant criterion is developed. Specifically, an inter-class penalty supervised by ground-truth labels is built on the source domain; this penalty aims to construct a separable structure for classes. Then a probabilistic and truncated intra-class agreement is proposed on the target domain, which treats the classes of the source domain as anchors and acquires the inter-class separability transductively.
\item Based on the above criterion, a global approximation scheme is extended. To capture the global structure, it combines the global information in the last epoch with data in the current batch. Since such approximation only requires access to the class-wise centers, it is actually memory-saving.
\item The manifold alignment is developed to be compatible with the embedding discriminant space. It establishes a series of abstract descriptors (i.e. the basis) for original data, and aligns the domains by minimizing the discrepancy between the abstract descriptors, while most of noise are filtered. Further, a theoretical error bound is derived to facilitate the selection of components.
\end{itemize}
\section{Related Work}\label{sec2}
Traditional UDA models usually focus on learning domain-invariant and discriminative features \cite{pan2010domain,long2013transfer}. Based on the manifold assumption, plentiful metrics are developed to measure the distance between instances from source and target \cite{gong2012geodesic,fernando2013unsupervised}.
Deep learning methods enhance the transferability by exploring the representations that disentangle exploratory factors of variants hidden behind the data \cite{bengio2013representation,yosinski2014transferable}. The distribution alignment methods minimize the discrepancy of domains based on common statistics directly, e.g., the first-order statistic based on maximum mean discrepancies (MMD) \cite{sejdinovic2013equivalence,long2015learning,ren2019learning} and the second-order statistic based on covariance matrices \cite{sun2016return,chen2019joint}. Inspired by the GANs \cite{goodfellow2014generative}, lots of adversarial approaches with different purposes are developed. The most common usage of adversarial networks is to generate the representations that fool the domain discriminator, thus the distributions of domains are more similar \cite{ganin2016domain,Long2017Deep,pinheiro2018unsupervised}. Domain-specific and Task-specific methods aim to tackle the issue of compact representations in high-level layers \cite{Long2017Deep,saito2018maximum,kim2019unsupervised,lee2019sliced,ding2019deep}.

Though adversarial alignment generates well marginal distributions, the conditional distributions still need to be explored. Recent researches suggest that discriminability plays a crucial role in the formation of class distributions (i.e., the conditional distributions) \cite{long2018conditional,ding2019deep,chen2019transferability}. Conditional Domain Adversarial Network (CDAN) \cite{long2018conditional} encodes the target predictions into deep features and then models the joint distributions of features and labels. Batch Spectral Penalization (BSP) \cite{chen2019transferability} revisits the relation between transferability and discriminability via the largest singular value of batch features.

\begin{figure*}[t]
\begin{center}
\includegraphics[width=0.95\linewidth]{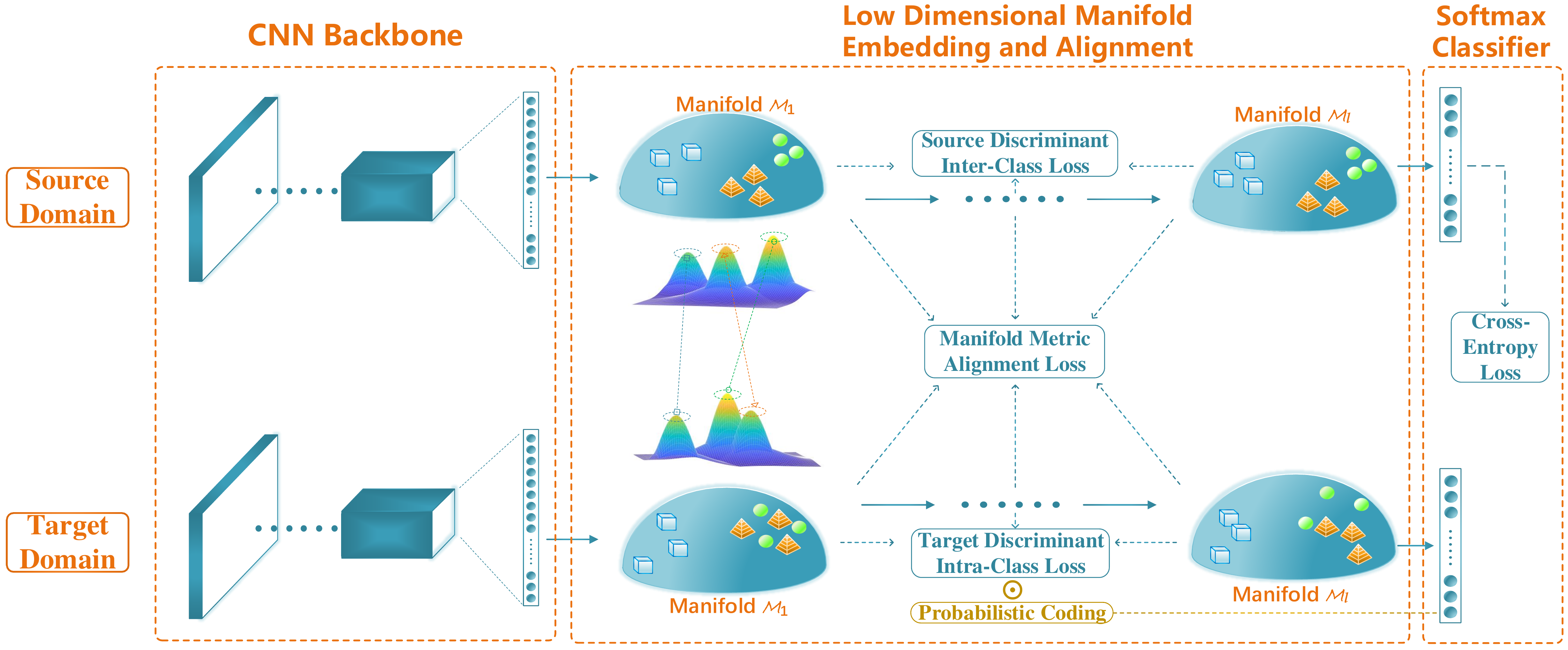}
\end{center}
   \caption{Overview of the proposed multilayer Riemannian manifolds embedding and alignment network. Stage 1: deep features based on CNNs. Stage 2: Riemannian manifold layers, where fully connected layers with proposed ``weak'' discriminant criterion and manifold metric domain alignment are employed to transfer the discriminative information.
   }
\label{fig:network}
\end{figure*}
\section{Multi-layer Remannian Manifold Embedding and Alignment}\label{sec3}
In this section, we propose the Discriminative Remannian Manifold Embedding and Alignment (DRMEA) framework.

\subsection{Backgrounds and Motivations}\label{sec3.1}
In the classical manifold learning paradigm, to construct a compact and discriminative embedding space, a low-dimensional manifold is usually extracted from the originally high-dimensional data space. Specifically, the Riemannian manifold $\mathcal{M}$ usually consists of a certain object such as linear subspace, affine/convex hull, symmetric positive definite (SPD) matrix \cite{huang2017cross}.

From the perspective of discriminative embedding, graph-based criterion \cite{yan2007graph} is widely adopted in the area of manifold learning and domain adaptation. Basically, those methods establish the instances-based connection graph or similarity graph to construct a separable space. Besides, as the primary assumption of domain adaptation is based on statistical distribution, the alignment based on covariance matrices, which lie on the Riemannian manifold, equips the domain with the manifold and statistical properties. Motivated by it, our work aims to embed the graph-based discriminant criterion to the target domain, which is represented as manifolds (i.e., the covariance matrices).

Given features ${\bf X} \in \mathbb{R}^{d\times n}$ and its mean vector $\bar{{\bf x}}\in\mathbb{R}^{d}$, where $d$ denote the dimension of features and $n$ represent the sample sizes. Denote by $S$ the input space (e.g., Euclidean space, Hilbert space or Manifold space), the manifold learning aims to learning a specific nonlinear mapping
\begin{equation*}
\small
f:~~ S ~ \rightarrow ~ \mathcal{M},
\end{equation*}
where $\mathcal{M}$ is the low-dimensional embedding manifold. Based on the SPD representation setting, the image of a given covariance matrix ${\bf C(X)}=\frac{1}{n-1} ({\bf X}-\bar{{\bf x}}{\bf 1}^T_n)({\bf X}-\bar{{\bf x}}{\bf 1}^T_n)^T \in\mathbb{R}^{d\times d}$ is a low-dimensional SPD matrix ${\bf C}' = f({\bf C}) \in\mathbb{R}^{d'\times d'}$, where ${\bf 1}_n$ is $n$-dimensional vector with all one elements and $(\cdot)^T$ is the transpose operation. Intuitively, learning of mapping function $f$ can be deduced to find a nonlinear transformation
\begin{equation*}
\small
g:~ {\bf X} ~ \mapsto g({\bf X})
\end{equation*}
and the image of mapping $f$ can be approximated by the inner product of $g({\bf X})$, i.e., $f({\bf C}) \approx g({\bf X})g({\bf X})^{T}$.

For domain adaptation, the source and target domains can be taken as two Euclidean spaces, where the discriminative information is relatively inadequate. Thus the ideal manifolds are expected to be discriminative, representative and compact. Besides, the features distribution of domains, which is represented by manifolds, should be aligned with manifold metric for the better transfer of discriminative structure.

\subsection{Low-Dimensional Manifold Layers}\label{sec3.2}
As previously stated, we aim to learn a nonlinear transformation $g$ for the input features ${\bf X}$ directly. In this paper, CNNs are used to obtain such projection $g$. To explore the latent Riemannian representations of the Euclidean features (i.e., the deep features in stage 1), the output features of CNN backbone are sent into progressive low-dimensional manifold layers in the second stage. Since there is a naturally geometric difference between Euclidean Space and Riemannian Space, a multilayer scheme is adopted to reduce the dimension of features progressively.

Figure \ref{fig:network} shows the network architecture of the proposed method. Let $\Theta$ be the parameters of networks. The progressive Riemannian manifold layers $\{\mathcal{M}_i|i=1,2,\ldots,l\}$ are represented as a sequence of functions $\{g_i|i=1,2,\ldots,l\}$, and implemented on fully connection layers. In fact, the CNNs and Riemannian manifold layers are generalized and share by both two domains. It means that the common projections are explored to map two domains to a general low-dimensional space. Therefore, any manifold layers $\mathcal{M}_i$ should be equipped with the following properties:
\begin{itemize}
\item Discriminative Structure: To strengthen the discriminative power of manifold space, the intra-class samples are required to be compact, while the inter-class samples are separable, respectively.
\item Consistent Structure: The source and target domains are aligned with manifold metric to match the manifold assumption. As a result, the domain discrepancy is represented as the distance between two submanifolds on $\mathcal{M}_i$, and then minimized based on the defined manifold metric (e.g., Grassmannian representations metric, Log-Euclidean metric and manifold principal angle similarity).
\end{itemize}
To reach the above goals, we propose to model the properties by losses $\mathcal{L}_{DS}$ and $\mathcal{L}_{AL}$, which will be detailed later. Then, the objective is formulated as following:
\begin{equation*}
\small
\mathop {\min}\limits_{\Theta} \mathcal{L} = \mathcal{L}_{CE} + \lambda_1 \mathcal{L}_{DS}  + \lambda_2 \mathcal{L}_{AL},
\end{equation*}
where $\mathcal{L}_{CE}$ is the cross-entropy loss of classifier on source domain and \{$\lambda_1,\lambda_2$\} are the penalty parameters.

\subsection{Discriminative Structure Loss}\label{sec3.3}
In this section, we describe how to embed the discriminative structure into the manifold layers. The main idea is shown in Figure \ref{fig:discriminant}. Since there exists a distribution discrepancy between different domains (e.g., (a) in Figure \ref{fig:discriminant}), conventional discriminant criterion is too strong to satisfy in this case. To relax the constraint, our method only focuses on the inter-class separability of the source domain and the intra-class compactness of the target domain.

Without loss of generality, we only introduce the formulation of the loss terms in $l$-th Riemannian manifold layer $\mathcal{M}_l$. Let ${\bf H}^s_l\in \mathbb{R}^{d_l\times n_s}$ and ${\bf H}^t_{l}\in \mathbb{R}^{d_l\times n_t}$ be the feature matrices of $\mathcal{M}_l$. Since class centers of the source domain are used in both two loss terms, the source mean vector ${\bf \bar{h}}^s_l\in\mathbb{R}^{d_l}$ and source class-wise mean matrix ${\bf \bar{H}}^s_l\in\mathbb{R}^{d_l\times c}$ are computed, where $c$ is the number of classes.

\subsubsection{Source Inter-Class Similarity}
\begin{figure}[t]
\begin{center}
\includegraphics[width=1\linewidth]{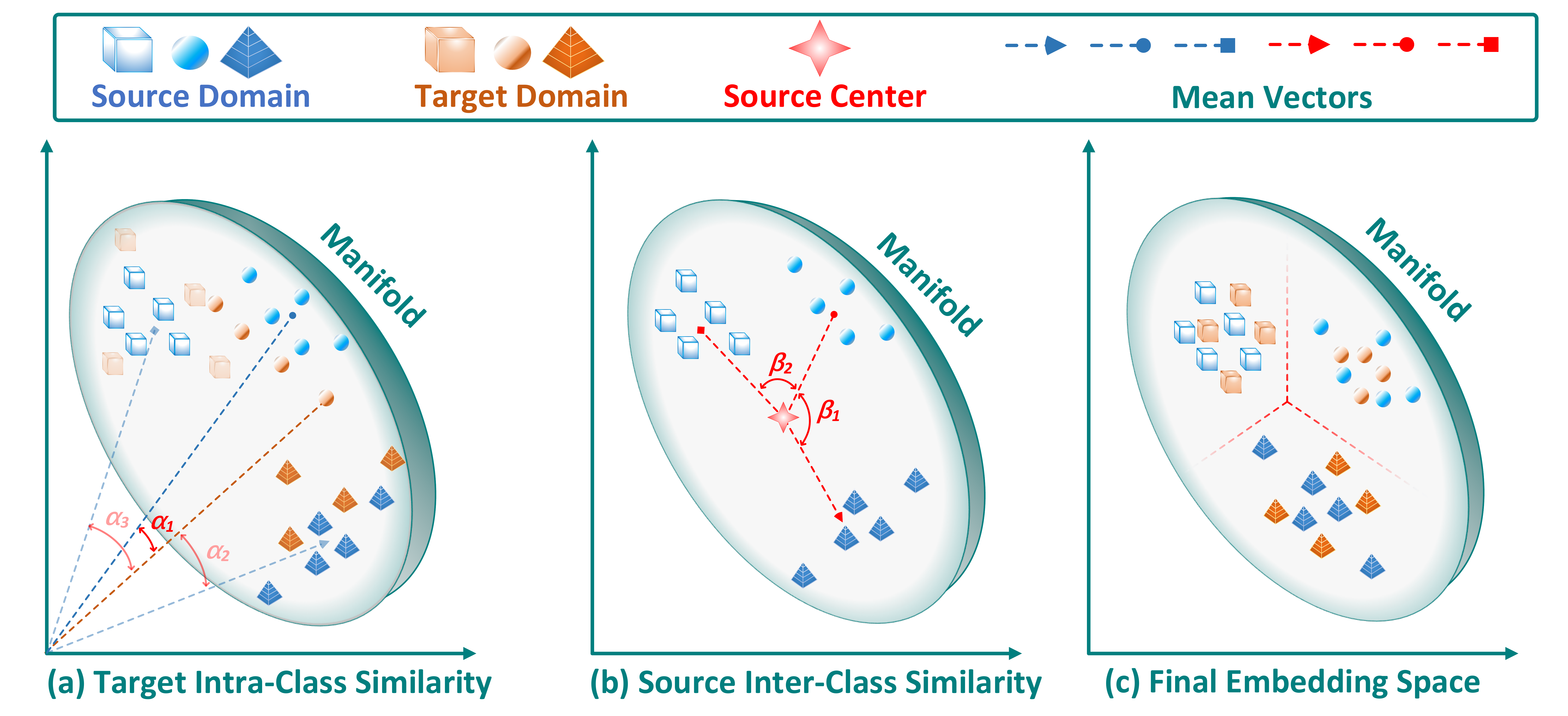}
\end{center}
   \caption{Illustration of the discriminative structure loss. 
   (a) The target intra-class similarity constructs a compact space for the samples from same category. (b) The source inter-class similarity forms a separable space for source samples by finding a optimal rotations. (c) The final embedding space, where the target domain is discriminative.}
\label{fig:discriminant}
\end{figure}

Though the traditional inter-class discriminant criterion is applicable on the source domain, a nice geometric structure of the class distribution is actually not guaranteed under the distance metric. To this end, the similarity measurement is utilized here, which has also shown in Figure \ref{fig:discriminant} (b). Rather than compute the similarities between class-wise centers and total center directly, we process the class-wise centers as following
\begin{equation*}
\small
{\bf \hat{H}}^s_l \triangleq  {\bf \bar{H}}^s_l - {\bf \bar{h}}^s_l{\bf 1}_c^T.
\end{equation*}
We call ${\bf \hat{H}}^s_l=[{\bf \hat{h}}^s_1,{\bf \hat{h}}^s_2,\ldots,{\bf \hat{h}}^s_c]$ the centralized class-wise means hereinafter. Further, if the columns of ${\bf \hat{H}}^s_l$ are normalized with $\ell_2$ norm, the cosine similarity matrix is derived as ${\bf S}^l_{inter} = {\bf \hat{H}}^{s^T}_l{\bf \hat{H}}^s_l$.
Because ${\bf S}^l_{inter}(i,j)={\bf \hat{h}}^{s^T}_i {\bf \hat{h}}^{s}_j$ indicates the similarity between $i$-th class and $j$-th class, the diagonal elements are meaningless. Then the separable structure is reached by maximizing the dissimilarities between the centralized class-wise mean vectors. Equivalently, it can be achieved by minimizing the following inter-class loss:
\begin{equation}\label{eq:1}
\small
\mathcal{L}^l_{inter}({\bf H}^{s}_l) = \frac{2}{c(c-1)}\sum_{i<j} {\bf S}^l_{inter}(i,j).
\end{equation}

Let us take Figure \ref{fig:discriminant} (b) as an example. There is a 2-dimensional space with 3 classes. Let \{1,2,3\} denotes the labels of ``Ball'', ``Pyramid'' and ``Cube'', respectively. Under this situation, ${\bf S}^l_{inter}(1,2)$ and ${\bf S}^l_{inter}(1,3)$ are depicted as $\cos(\beta_1)$ and $\cos(\beta_2)$, respectively. According to the goal of Eq. \eqref{eq:1} and ignoring the constraints, the optimal solution occurs at $\beta_1=\beta_2=\frac{2}{3}\pi$, and the minimal $\mathcal{L}^l_{inter}$ equals to $-\frac{1}{2}$ (which can also be seen as the lower bound of constrained scenarios).

\subsubsection{Target Intra-Class Similarity}
On the other hand, since there are no labels on the target domain, the discriminant learning is facilitated by the soft labels (i.e., the output of softmax layer). Let ${\bf P}^t = [{\bf p}^t_1,{\bf p}^t_2,\ldots,{\bf p}^t_{n_t}] \in \mathbb{R}^{c\times n_t}$ be the softmax predictions of classifier layer. Since ${\bf P}^t$ can be regarded as the confidence or probability of classification, the predictions are used to weight the importance or confidence of the supervised information provided by soft labels. Similarly, assuming the columns of ${\bf \bar{H}}^s_l$ and ${\bf H}^t_{l}$ have unit length. The similarities under all classification cases can be written as ${\bf S}^l_{intra} = {\bf \bar{H}}^{s^T}_l {\bf H}^t_l$. It means that the source class-wise centers are utilized instead of the target. The main reasons can be summarized as follows: the inter-class structure learned from the source domain can be transduced to the target domain; the source class-wise centers computed from ground-truth labels are more reliable. Because there is so much uncertainty when pseudo labels are straightforwardly used on the target domain, we establish a probabilistic discriminative criterion to make the most of the information provided by soft labels. Intuitively, ${\bf P}^t$ is a natural choice for the probabilistically weighting model. Then the probabilistic intra-class loss is formalized as
\begin{equation}\label{eq:2}
\small
\mathcal{L}^l_{intra}({\bf H}^t_{l},{\bf P}^t) = -\frac{1}{n_t c}\sum_{i=1}^{c} \sum_{j=1}^{n_t} {\bf P}^t(i,j) {\bf S}^l_{intra}(i,j).
\end{equation}

However, there are much noise in ${\bf P}^t$, whose values are very small. Especially when the softmax classifier comes to converging, the columns of ${\bf P}^t$ tend to be the one-hot vectors. As truncation is a efficient way for denoising, we develop a Top-$k$ preserving scheme for the truncated intra-class loss. Let $V_j = \{(i,j)|i=v_{1j},v_{2j},\ldots,v_{kj}\}$ be the index set of $k$-largest elements in ${\bf p}^t_j$, $j=1,2,\ldots,n_t$. Then a characteristic function like matrix is defined as
\begin{equation*}
\small
\scalebox{1.3}{$\chi$} (i,j) = \left\{
\begin{aligned}
1,&~~ (i,j)\in V_j, \\
0,&~~ (i,j)\notin V_j.
\end{aligned}
\right.
\end{equation*}
Then, the intra-class loss is modified by the truncated matrix \scalebox{1.3}{$\chi$} and written as
\begin{equation}\label{eq:3}
\small
\mathcal{L}^l_{intra}({\bf H}^t_{l},{\bf P}^t) = -\frac{1}{n_t k}\sum_{i=1}^{c} \sum_{j=1}^{n_t} \scalebox{1.3}{$\chi$}(i,j) {\bf P}^t(i,j) {\bf S}^l_{intra}(i,j).
\end{equation}

A simple illustration is also shown in Figure \ref{fig:discriminant} (a). Based on the previous notations, ${\bf S}^l_{intra}(1,1)$, ${\bf S}^l_{intra}(1,2)$, ${\bf S}^l_{intra}(1,3)$ are computed as $\cos(\alpha_1)$, $\cos(\alpha_2)$ and $\cos(\alpha_3)$, respectively. Suppose the softmax output of the ``Ball'' sample in figure is ${\bf p}^t_1 = [0.75,0.15,0.1]^T$. According to Eq. \eqref{eq:2}, all three similarities are taken into consideration, while $\cos(\alpha_2)$ and $\cos(\alpha_3)$ are noise. If we adopt the Top-$2$ strategy in Eq. \eqref{eq:3}, the perturbation from the ``Cube'' $\cos(\alpha_3)$ can be excluded.

In conclusion, the proposed two loss terms build a probabilistic discriminant criterion on the target domain. The ground-truth labels on the source domain provide a reliable separable structure directly, where the intra-class structure is unnecessary. Then the target samples are attached to the corresponding source class-wise center via soft labels. As shown in Figure \ref{fig:discriminant} (c), the intra-class relationship on the source domain does not change much while the discriminative property of the target domain is satisfied. Finally, the discriminative structure loss is noted by
\begin{equation*}
\mathcal{L}_{DS} = \sum_{i} (\mathcal{L}^i_{inter} + \mathcal{L}^i_{intra}),
\end{equation*}

\subsubsection{Global Structure Learning}
For the batch scheme in deep models, it is hard to obtain the complete relation graph between the instances. The direct application of classical graph embedding may be misled by some extreme local distributions, which will result in a suboptimal solution.

Supposing that the geometry of manifolds does not change drastically after several updates, we can built some \textit{anchors} in the whole data space to acquire the global information. In this work, we propose to fix the \textit{anchors} in each batch iteration and update them after every epoch. Specifically, the \textit{anchors}, i.e., ${\bf \bar{h}}^s_l$ in inter-class loss Eq. \eqref{eq:1} and ${\bf \bar{H}}^{s}_l$ in intra-class loss Eq. \eqref{eq:3}, are computed from the last epoch. Note that the \textit{anchors} are treated as constants in optimization. ${\bf \bar{H}}^{s}_l$ in Eq. \eqref{eq:1} and ${\bf H}^t_l$ in Eq. \eqref{eq:3} are obtained from batch data. The inter-class loss strongly supervised by source labels is imposed at the beginning. While the intra-class loss facilitated by soft label is equipped after a certain number of iterations/epoches.

\subsection{Manifold Metric Alignment Loss}\label{sec3.4}
\begin{table*}[t]
\caption{Recognition Rates (\%) on VisDA-2017 (ResNet-101) and Office-Home (ResNet-50).}
\label{tab_VisDA2017&OfficeHome}
\begin{center}
\resizebox{2.1\columnwidth}{!}
{
\begin{tabular}{c|p{24pt}p{24pt}p{24pt}p{24pt}p{24pt}p{24pt}p{24pt}p{24pt}p{24pt}p{24pt}p{24pt}p{28pt}|c}
\toprule[1pt]
\textbf{VisDA-2017} & Plane & bcycl & bus & car & horse & knife & mcyle & person & plant & sktbrd & train & truck & Mean \\
\hline
ResNet-101 \cite{he2016deep} & 55.1 &  53.3 &  61.9 &  59.1 &  80.6 &  17.9 &  79.7 &  31.2 &  81.0 &  26.5 &  73.5 &  8.5 &  52.4 \\
DAN \cite{long2015learning} & 87.1 &  63.0 &  76.5 &  42.0 &  90.3 &  42.9 &  85.9 &  53.1 &  49.7 &  36.3 &  85.8 &  20.7 &  61.1 \\				
DANN \cite{ganin2016domain} & 81.9 &  77.7 &  82.8 &  44.3 &  81.2 &  29.5 &  65.1 &  28.6 &  51.9 &  54.6 &  82.8 &  7.8 &  57.4 \\
MCD \cite{saito2018maximum} & 87.0 &  60.9 &  83.7 &  64.0 &  88.9 &  79.6 &  84.7 &  76.9 &  88.6 &  40.3 &  83.0 &  25.8 &  71.9 \\
SimNet \cite{pinheiro2018unsupervised} & \textbf{94.3} & \textbf{ 82.3} &  73.5 &  47.2 &  87.9 &  49.2 &  75.1 &  \textbf{79.7} & 85.3 &  68.5 &  81.1 &  \textbf{50.3} & 72.9 \\
GTA \cite{sankaranarayanan2018generate} & - & - &  - &  - & - & - & - & - & - &  - &  - &  - & 77.1 \\
CDAN \cite{long2018conditional} & 85.2 &  66.9 &  83.0 &  50.8 &  84.2 &  74.9 &  88.1 &  74.5 &  83.4 &  76.0 &  81.9 &  38.0 &  73.7 \\
GPDA \cite{kim2019unsupervised} & 83.0 &  74.3 &  80.4 &  66.0 &  87.6 &  75.3 &  83.8 &  73.1 &  90.1 &  57.3 &  80.2 &  37.9 &  73.3 \\
BSP+DANN \cite{chen2019transferability} & 92.2 &  72.5 &  83.8 &  47.5 &  87.0 &  54.0 &  86.8 &  72.4 &  80.6 &  66.9 &  84.5 &  37.1 &  72.1 \\
BSP+CDAN \cite{chen2019transferability} & 92.4 &  61.0 &  81.0 &  57.5 &  89.0 &  80.6 &  90.1 &  77.0 &  84.2 &  \textbf{77.9} & 82.1 &  38.4 &  75.9 \\
\hline
DRMEA (No AL) & 92.8 & 15.3 & \textbf{86.7} & \textbf{86.3} & \textbf{93.8} & 70.7 & \textbf{95.2} & 68.9 & \textbf{95.8} & 40.4 & 85.1 & 5.6 & 69.7 \\
DRMEA (No DS) & 90.2 & 66.5 & 70.2 & 65.8 & 79.8 & 81.8 & 84.7 & 70.1 & 82.0 & 46.5 & \textbf{88.1} & 27.7 & 71.1 \\
DRMEA & 92.1 & 75.0 & 78.9 & 75.5 & 91.2 & \textbf{81.9} & 89.0 & 77.2 & 93.3 & 77.4 & 84.8 & 35.1 & \textbf{79.3} \\
\bottomrule[1pt]
\toprule[1pt]
\textbf{Office-Home} & Ar$\rightarrow$Cl & Ar$\rightarrow$Pr & Ar$\rightarrow$Rw & Cl$\rightarrow$Ar & Cl$\rightarrow$Pr & Cl$\rightarrow$Rw &
Pr$\rightarrow$Ar & Pr$\rightarrow$Cl & Pr$\rightarrow$Rw & Rw$\rightarrow$Ar & Rw$\rightarrow$Cl & Rw$\rightarrow$Pr & Mean \\
\hline
ResNet-50 \cite{he2016deep} & 34.9 & 50.0 & 58.0 & 37.4 & 41.9 & 46.2 & 38.5 & 31.2 & 60.4 & 53.9 & 41.2 & 59.9 & 46.1 \\
DAN \cite{long2015learning} & 43.6 & 57.0 & 67.9 & 45.8 & 56.5 & 60.4 & 44.0 & 43.6 & 67.7 & 63.1 & 51.5 & 74.3 & 56.3 \\				
DANN \cite{ganin2016domain} & 45.6 & 59.3 & 70.1 & 47.0 & 58.5 & 60.9 & 46.1 & 43.7 & 68.5 & 63.2 & 51.8 & 76.8 & 57.6 \\
JAN \cite{Long2017Deep} & 45.9 & 61.2 & 68.9 & 50.4 & 59.7 & 61.0 & 45.8 & 43.4 & 70.3 & 63.9 & 52.4 & 76.8 & 58.3 \\
CDAN \cite{long2018conditional} & 49.0 & 69.3 & 74.5 & 54.4 & 66.0 & 68.4 & 55.6 & 48.3 & 75.9 & 68.4 & 55.4 & 80.5 & 63.8 \\
CDAN+E \cite{long2018conditional} & 50.7 & 70.6 & 76.0 & 57.6 & 70.0 & 70.0 & 57.4 & 50.9 & 77.3 & 70.9 & 56.7 & 81.6 & 65.8 \\
BSP+DANN \cite{chen2019transferability} & 51.4 & 68.3 & 75.9 & 56.0 & 67.8 & 68.8 & 57.0 & 49.6 & 75.8 & 70.4 & 57.1 & 80.6 & 64.9 \\
BSP+CDAN \cite{chen2019transferability} & 52.0 & 68.6 & 76.1 & 58.0 & 70.3 & 70.2 & 58.6 & 50.2 & 77.6 & \textbf{72.2} & \textbf{59.3} & \textbf{81.9} & 66.3 \\
\hline
DRMEA (No AL) & 51.9 & 72.8 & 77.1 & 63.0 & \textbf{72.0} & 71.3 & 60.5 & 49.5 & 78.4 & 71.5 & 54.4 & 82.8 & 67.1 \\
DRMEA (No DS) & 51.2 & 72.4 & \textbf{77.7} & 63.0 & 71.4 & 71.4 & 58.6 & 44.6 & \textbf{79.1} & 71.1 & 53.4 & 81.5 & 66.3 \\
DRMEA & \textbf{52.3} & \textbf{73.0} & 77.3 & \textbf{64.3} & \textbf{72.0} & \textbf{71.8} & \textbf{63.6} & \textbf{52.7} & 78.5 & 72.0 & 57.7 & 81.6 & \textbf{68.1} \\
 & $\pm$0.4 & $\pm$0.6 & $\pm$0.3 & $\pm$0.3 & $\pm$0.7 & $\pm$0.5 & $\pm$0.6 & $\pm$0.7 & $\pm$0.2 & $\pm$0.1 & $\pm$0.6 & $\pm$0.2 & $\pm$0.2 \\
\bottomrule[1pt]
\end{tabular}
}
\end{center}
\end{table*}

To satisfy the second property, i.e., Consistent Structure, a manifold metric alignment method is developed. As mentioned before, the covariance matrix is an important tool to represent a manifold $\mathcal{M}$. Therefore, the alignment based on covariance not only meets the requirements of manifold metric, but also reaches some nice statistical properties, such as distribution assumption.

\subsubsection{Grassmannian Metric}\label{sec3.4.1:grass_dist}
Let ${\bf C}^s_l$ and ${\bf C}^t_l$ be the covariance matrices of source and target domains computed from batch-wise features, respectively. Assume $\mathcal{M}^s_l$ and $\mathcal{M}^t_l$ are two submanifolds of $\mathcal{M}_l$, which are represented by their corresponding covariance matrices. Before the alignment process, these two submanifolds are partially overlapped, and our goal is to minimize the discrepancy under the metric of $\mathcal{M}_l$. In general, the manifold metric alignment loss of the $l$-th layer is expressed as
\begin{equation}\label{eq:4}
\small
\mathcal{L}_{align}^{l} \triangleq dist(\mathcal{M}^s_l,\mathcal{M}^t_l) = d_{\mathcal{M}}({\bf C}^s_l,{\bf C}^t_l),
\end{equation}
where $d_{\mathcal{M}}(\cdot,\cdot)$ is the manifold metric to be determined.

Grassmannian manifold is a well-known type of Riemannian manifold. It is a projection subspace $\mathbb{R}^{d'_l}$ deduced from the originally high-dimensional space $\mathbb{R}^{d_l}$, $d'_l<d_l$. Thus the two submanifolds $\mathcal{M}^s_l$ and $\mathcal{M}^t_l$ lying on the Grassmannian manifold $\mathcal{M}_l$ are represented as two individual points. The distance between such two points is measured by the discrepancy between their projection orthogonal basis ${\bf U}^s_l$ and ${\bf U}^t_l$. Specifically, the orthogonal basis of such $d'_l$-dimensional Grassmannian manifold consists of $d'_l$ dominant singular vectors with respect to its representation matrix. Thus ${\bf U}^s_l$ and ${\bf U}^t_l$ are two $d_l\times d'_l$ column-orthogonal matrices, which can be obtained from the Singular Value Decomposition (SVD) of covariance matrices ${\bf C}^s_l$ and ${\bf C}^t_l$, respectively. Finally, the Grassmannian distance is measured by
\begin{equation}\label{eq:5}
\small
d_{\mathcal{M}}({\bf C}^s_l,{\bf C}^t_l) = \frac{1}{d^2_l}\| {\bf U}^s_l {\bf U}^{s^T}_l - {\bf U}^t_l {\bf U}^{t^T}_l \|^2_F,
\end{equation}
where $\| \cdot \|_F$ is the Frobenius norm. Thus the manifold metric alignment loss can be written as
\begin{equation*}
\small
\mathcal{L}_{AL} = \sum_{i} \mathcal{L}_{align}^{i},
\end{equation*}

\subsubsection{Error Bound of Grassmannian Metric}\label{sec3.4.1:grass_gheory}
As the dimension $d'_l$ is needed in Grassmannian distance, we establish an theoretical error bound for it. Inspired by the previous works \cite{zwald2006convergence,fernando2013unsupervised}, we shall denote the covariance of given distribution $D$ by ${\bf C}$, and covariance drawn i.i.d. from $D$ with sample size $n$ by $\tilde{\bf C}$. Then Zwald et al. \cite{zwald2006convergence} give the following theorem.
\begin{theorem}\cite{zwald2006convergence}\label{thm:1}
Let B be s.t. for any vector ${\bf x}$, $\| {\bf x} \| \leq B$, let ${\bf U}^{d'}_{{\bf C}}$ and ${\bf U}^{d'}_{\tilde{{\bf C}}}$ be the orthogonal projectors of the subspaces spanned by the first $d'$ eigenvectors of ${\bf C}$ and $\tilde{\bf C}$, respectively. Let $\lambda_1>\lambda_2>\cdots>\lambda_{d'}>\lambda_{d'+1}\geq0$ be the first $d'+1$ eigenvalues of ${\bf C}$, then for any
$n \geq\left(\frac{4 B}{\lambda_{d'}-\lambda_{d'+1}}\left(1+\sqrt{\frac{\ln (1 / \delta)}{2}}\right)\right)^{2}$
with probability at least $1-\delta$ we have:
\begin{equation}\label{eq:6}
\small
\|{\bf U}^{d'}_{{\bf C}}-{\bf U}^{d'}_{\tilde{{\bf C}}}\| \leq \frac{4 B}{\sqrt{n}\left(\lambda_{d'}-\lambda_{d'+1}\right)}\left(1+\sqrt{\frac{\ln (1 / \delta)}{2}}\right).
\end{equation}
\end{theorem}
Above theorem shows the relation between the error and $d'$. Defining the right side of Eq. \eqref{eq:6} as $\frac{E(\delta)}{\lambda_{d'}-\lambda_{d'+1}}$. To extend the inequality to the Grassmannian distance, we derive following lemma.
\begin{lemma}\label{lem:2}
Based on the condition in Theorem \ref{thm:1}, we have
\begin{equation*}
\small
\|{\bf U}^{d'}_{{\bf C}}{\bf U}^{d'^T}_{{\bf C}}-{\bf U}^{d'}_{\tilde{{\bf C}}}{\bf U}^{d'^T}_{\tilde{{\bf C}}}\|_F \leq 2\sqrt{2}E(\delta) \frac{\sqrt{d'}}{\lambda_{d'}-\lambda_{d'+1}}
\end{equation*}
with probability at least $1-\delta$.
\end{lemma}

Based on Lemma \ref{lem:2}, following theorem gives the error of $d_{\mathcal{M}}({\bf C}^s,{\bf C}^t)$ with respect to its $n$ samples approximation $d_{\tilde{\mathcal{M}}}(\tilde{\bf C}^s,\tilde{\bf C}^t)$.
\begin{theorem}\label{thm:3}
Assuming the condition in Theorem \ref{thm:1} is specified by domains. Specifically, $\lambda_i^s$ and $\lambda_i^t$ denote the $i$-th largest eigenvalue of domain-specific covariance matrices ${\bf C}^s$ and ${\bf C}^t$, respectively. Denote by
\begin{equation*}
\small
e(d')=\frac{\sqrt{d'}}{\lambda_{d'}^s-\lambda_{d'+1}^s} + \frac{\sqrt{d'}}{\lambda_{d'}^t-\lambda_{d'+1}^t}
\end{equation*}
the error index. Then the following error bound holds with probability at least $1-\delta$:
\begin{equation*}
\small
|d_{\mathcal{M}}({\bf C}^s,{\bf C}^t)-d_{\tilde{\mathcal{M}}}(\tilde{\bf C}^s,\tilde{\bf C}^t)| \leq 2\sqrt{2}E(\delta)e(d').
\end{equation*}
\end{theorem}

Theorem \ref{thm:3} suggests that the upper bound of error is proportional to $e(d')$. It means that we should search the maximal gap between the continuous eigenvalues with the consideration of inflation factor $\sqrt{d'}$. Recall that in batch learning setting, the batch size $b_s$ is usually smaller than $d$, thus $d'$ only need to be searched in $\{1,2,\ldots,b_s-1\}$. The proofs are given in the Supplementary.
\section{Experiments and Comparative Analysis}\label{sec4}
In this section, three popular domain adaptation datasets are selected and the standard evaluation protocols are adopted.

\textbf{Office-Home} \cite{OfficeHome} contains 4 domains, i.e., \textit{Art} (\textbf{Ar}), \textit{Clipart} (\textbf{Cl}), \textit{Product} (\textbf{Pr}) and \textit{Real-World} (\textbf{Rw}).

\textbf{Image-CLEF-DA}\footnote{\url{https://www.imageclef.org/2014/adaptation}} consists of 4 domains. Following the previous protocol \cite{long2018conditional}, we conduct adaptation task between \textit{Caltech} (\textbf{C}), \textit{ImageNet} (\textbf{I}) and \textit{Pascal} (\textbf{P}).

\textbf{VisDA-2017} \cite{VisDA-2017} is a large-scale visual domain adaptation challenge dataset. The \textbf{synthetic} data to \textbf{real-image} track is evaluated here.

\subsection{Setup}
 Two layers Riemmanian manifold learning scheme is carried out in all experiments (i.e., $l=2$), where the first layer (1024d) is activated by Leaky ReLU ($\alpha = 0.2$) and the second layer (512d) by Tanh. Adam Optimizer ($lr=0.0002$, $\beta_1=0.9$, $\beta_2 = 0.999$) with batch size of 50 is utilized on Office-Home and Image-CLEF-DA datasets; the modified mini-batch SGD \cite{ganin2016domain} ($lr=0.003$, momentum $=$ 0.9, weight decay $=$ $5e-4$) with batch size of 32 is employed on VisDA-2017 challenge. The learning rate of CNN backbone layers is set as $0.1lr$. The hyperparameters are determined by try-and-error approach. Specifically, $\lambda_1$ and $\lambda_2$ are set as $1e1$ and $5e3$, respectively. The Top-$1$ scheme is adopted for the target intra-class loss in Eq. \eqref{eq:3}. For ablation study, the model without discriminative structure loss and manifold metric alignment loss are abbreviated as DRMEA (No DS) and DRMEA (No AL), respectively.

\subsection{Results Analysis}
\subsubsection{Error Bound of Grassmannian Distance}
\begin{figure}[t]
\begin{center}
\includegraphics[width=0.49\linewidth]{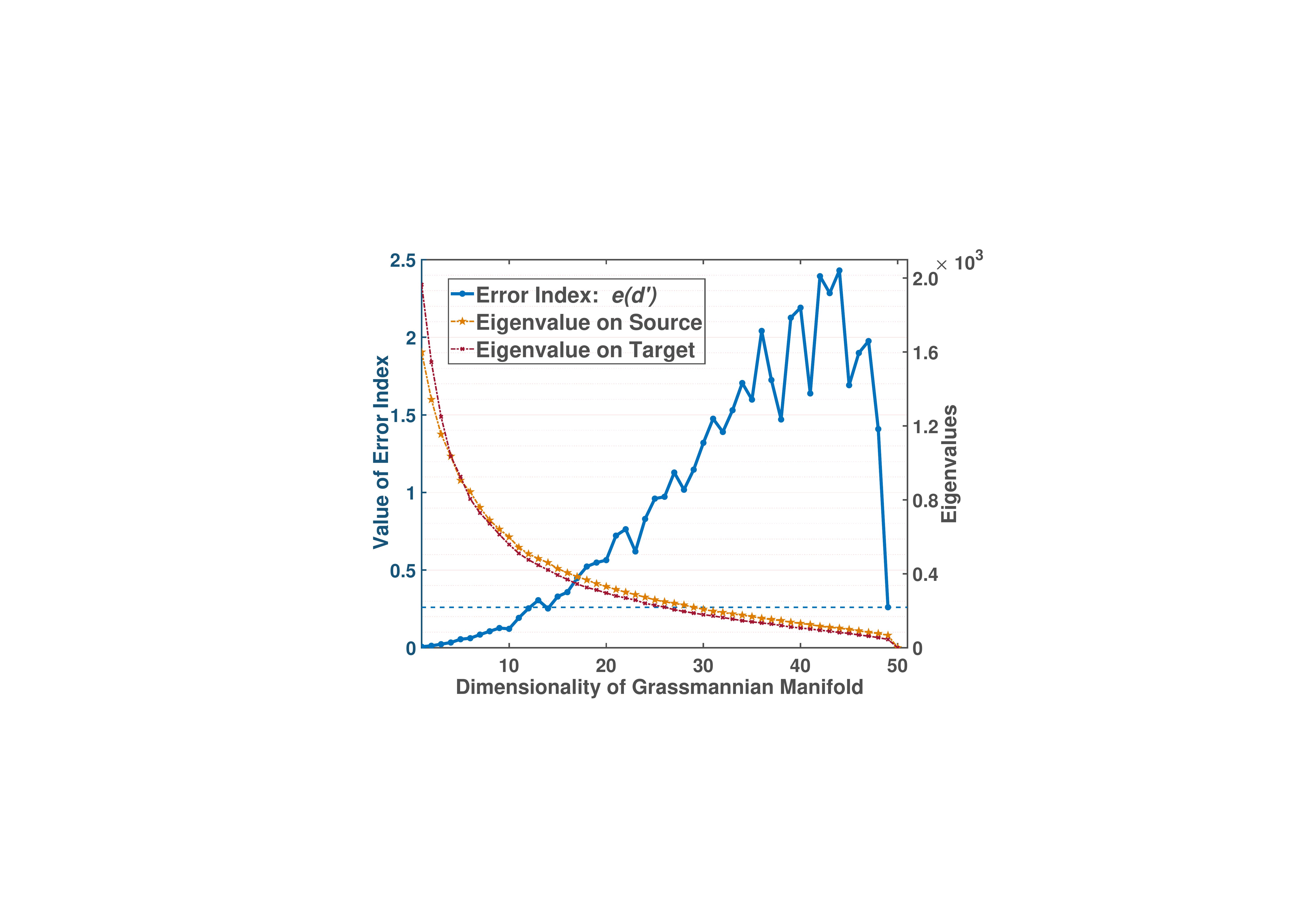}
\includegraphics[width=0.49\linewidth]{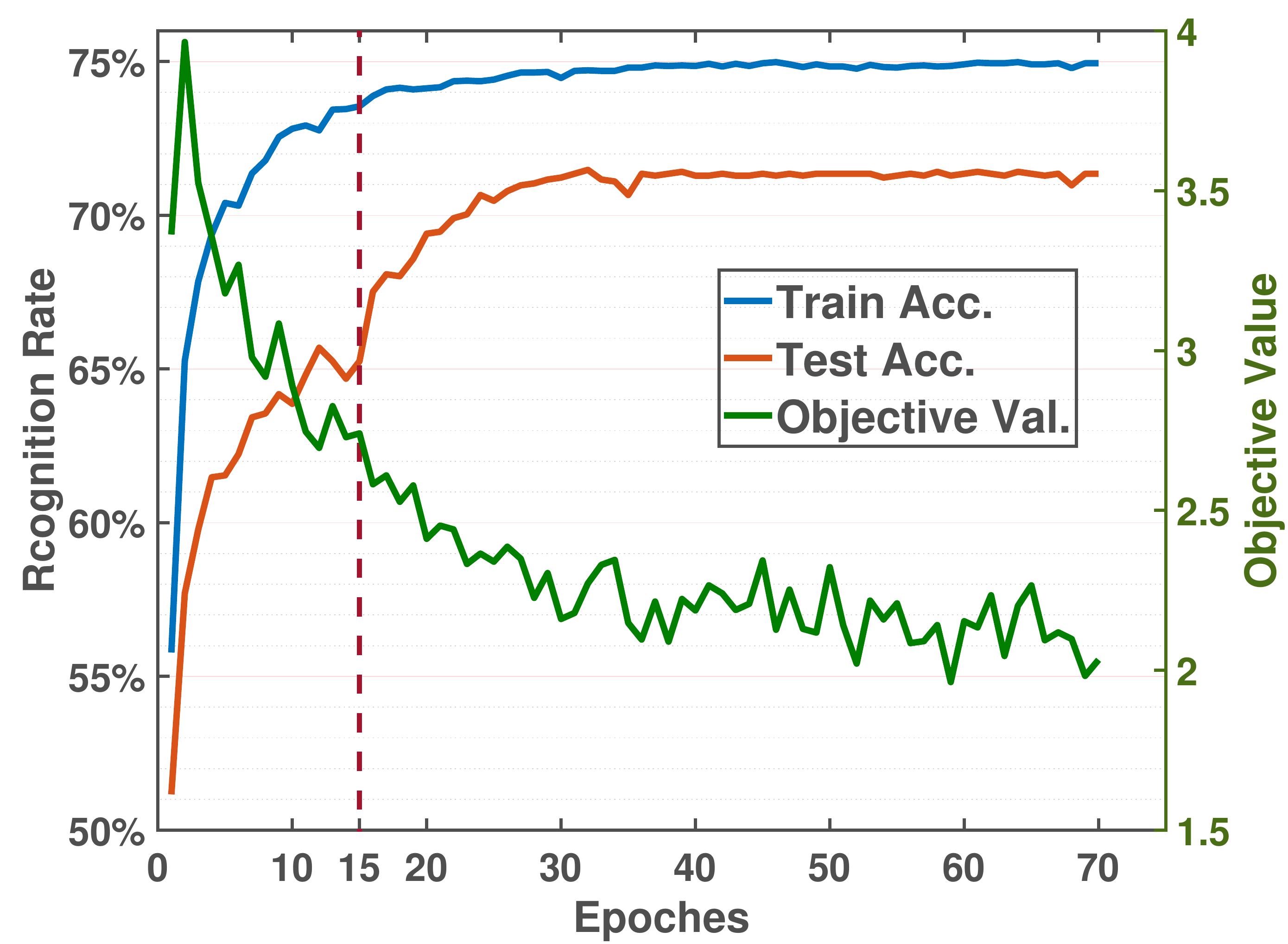}
\end{center}
   \caption{Left: Error and eigenvalue curves w.r.t. $d'$. Right: Recognition rate curves and loss curve.}
\label{fig:Error&Convergence}
\end{figure}

\begin{table*}[t]
\small
\caption{Recognition Rates (\%) on Image-CLEF-DA (ResNet-50).}
\label{tab_Image-CLEF-DA}
\begin{center}
\begin{tabular}{c|cccccc|c}
\toprule[1pt]
\textbf{Image-CLEF-DA} & I$\rightarrow$P & P$\rightarrow$I &
I$\rightarrow$C & C$\rightarrow$I &
C$\rightarrow$P & P$\rightarrow$C & Mean \\
\hline
ResNet-50 \cite{he2016deep} & 74.8 $\pm$ 0.3 & 83.9 $\pm$ 0.1 & 91.5 $\pm$ 0.3 & 78.0 $\pm$ 0.2 & 65.5 $\pm$ 0.3 & 91.2 $\pm$ 0.3 & 80.7 \\
DAN \cite{long2015learning} & 74.5 $\pm$ 0.4 & 82.2 $\pm$ 0.2 & 92.8 $\pm$ 0.2 & 86.3 $\pm$ 0.4 & 69.2 $\pm$ 0.4 & 89.8 $\pm$ 0.4 & 82.5 \\				
DANN \cite{ganin2016domain} & 75.0 $\pm$ 0.3 & 86.0 $\pm$ 0.3 & 96.2 $\pm$ 0.4 & 87.0 $\pm$ 0.5 & 74.3 $\pm$ 0.5 & 91.5 $\pm$ 0.6 & 85.0 \\
JAN \cite{Long2017Deep} & 76.8 $\pm$ 0.4 & 88.0 $\pm$ 0.2 & 94.7 $\pm$ 0.2 & 89.5 $\pm$ 0.3 & 74.2 $\pm$ 0.3 & 91.7 $\pm$ 0.3 & 85.8 \\
CDAN \cite{long2018conditional} & 76.7 $\pm$ 0.3 & 90.6 $\pm$ 0.3 & 97.0 $\pm$ 0.4 & 90.5 $\pm$ 0.4 & 74.5 $\pm$ 0.3 & 93.5 $\pm$ 0.4 & 87.1 \\
CDAN+E \cite{long2018conditional} & 77.7 $\pm$ 0.3 & 90.7 $\pm$ 0.2 & \textbf{97.7} $\pm$ 0.3 & \textbf{91.3} $\pm$ 0.3 & 74.2 $\pm$ 0.2 & 94.3 $\pm$ 0.3 & 87.7 \\
\hline
DRMEA (No AL) & 78.0 $\pm$ 0.1 & 91.1 $\pm$ 0.1 & 95.6 $\pm$ 0.2 & 88.7 $\pm$ 0.3 & 74.8 $\pm$ 0.1 & 94.8 $\pm$ 0.2 & 87.3 \\
DRMEA (No DS) & 78.9 $\pm$ 0.1 & 90.5 $\pm$ 0.2 & 94.0 $\pm$ 0.1 & 87.8 $\pm$ 0.1 & 76.7 $\pm$ 0.2 & 93.0 $\pm$ 0.1 & 86.8 \\
DRMEA & \textbf{80.7} $\pm$ 0.1 & \textbf{92.5} $\pm$ 0.1 & 97.2 $\pm$ 0.1 & 90.5 $\pm$ 0.1 & \textbf{77.7} $\pm$ 0.2 & \textbf{96.2} $\pm$ 0.2 & \textbf{89.1} \\
\bottomrule[1pt]
\end{tabular}
\end{center}
\end{table*}

The numerical simulation is conducted on Image-CLEF-DA dataset to explore the minimal error index $e(d')$. As a fact that the eigenvalues always decrease rapidly at the beginning and enter into a flat state, the error bounds of dimensionality $d'$ located in the flatten area are too high to assess. As shown in Figure \ref{fig:Error&Convergence}, the trend of eigenvalues is consistent with the description. Though the dramatic decrease in the beginning stage results in a lower error, the information in that area is unconvincing and insufficient to support the measurement of manifolds. Since there is a natural gap between the $(b_s-1)$-th and $b_s$-th dominant eigenvalues, $e(b_s-1)$ is smaller than most of other errors. We highlight the error index of $e(b_s-1)$ by blue dash line, and observe only errors of $d'=\{1,2,\ldots,12,14\}$ are lower than $e(b_s-1)$. Empirically, the dimensionality of Grassmannian manifold $d'$ is set as $b_s-1$ hereinafter.

\subsubsection{Convergence}
The convergence cures on Office-31 \textbf{A}$\rightarrow$\textbf{W} adaptation task are displayed in Figure \ref{fig:Error&Convergence}. It the beginning, the objective loss value decreases quickly and the recognition rate tends to enter a stable region in the epoch 10-15. However, the intra-class constraint is imposed after 15 epoches, which further activates the learning of discriminative structure. Thus the second ascent of accuracy on target domain occurred after 15 epoches, which leads to the continuous improvement of recognition rate and alleviates the over-fitting the on the source domain.

\begin{figure}[t]
\small
\begin{center}
\includegraphics[width=1\linewidth]{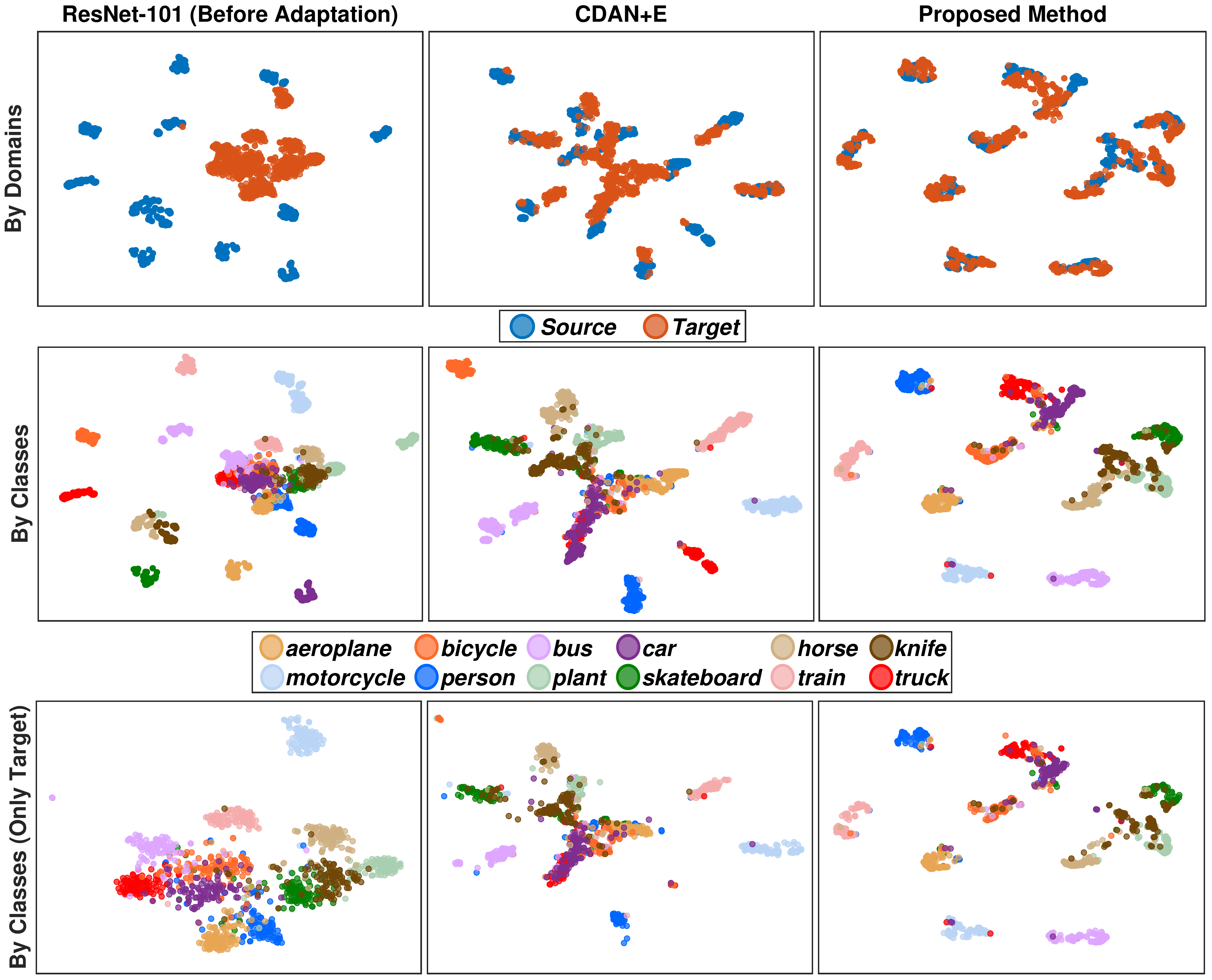}
\end{center}
   \caption{Visualization of the feature space on VisDA-2017 dataset. Rows are colored by domains or classes.
   }
\label{fig:tSNE}
\end{figure}
\subsubsection{Comparison}
Several state-of-the-art UDA approaches are selected and shown in Table \ref{tab_VisDA2017&OfficeHome}-\ref{tab_Image-CLEF-DA}. The experimental result on Visda-2017 dataset is shown in the top of Table \ref{tab_VisDA2017&OfficeHome}. We observe that DRMEA outperforms others by a large margin in average accuracy from the result. Performance on Office-Home dataset is provided at the bottom of Table \ref{tab_VisDA2017&OfficeHome}, the proposed method improves the accuracy to 68.1\% and obtains the highest accuracy in most of the adaptation tasks. Results on Image-CLEF-DA dataset are provided in Table \ref{tab_Image-CLEF-DA}. As the discrepancy between the source and target domains on Image-CLEF-DA dataset is relatively smaller than others, a more discriminative model is essential to the improvement of recognition accuracy. DRMEA encodes the discriminant criterion and alignment constraint simultaneously, thus it outperforms other methods by 1.4\% at least.

The ablation results also prove that the whole Riemannian manifold learning framework effect when both loss terms are equipped. As the discriminative structure loss provides a separable structure and manifold metric alignment loss bridges the distribution discrepancy between the source and target domains based on Grassmannian distance, both two losses are important.
\subsubsection{Visualization}
Figure \ref{fig:tSNE} shows the 2-D representation spaces obtained from t-SNE \cite{maaten2008visualizing} algorithm on VisDA-2017 dataset. CDAN+E shortens the distance between source and target by using adversarial alignment. A part of classes has been dragged away from the center, e.g., \textit{plant, car, horse, aeroplane} and \textit{bicycle}. In the third column, our method further optimizes the structure of the representation space. The categories are aligned better than ResNet-101 and CDAN+E, which leads to more compact target space.
\section{Conclusion}\label{sec5}
In this paper, we develop a Riemannian manifold embedding and alignment framework for UDA, where the transferability and discriminability are reached consistently. To optimize the structure of the target domain, the soft labels are encoded into the discriminant criterion probabilistically and transductively. Then a globally discriminative structure is approximated via a memory-saving manner. A theoretical error bound is derived, which is guaranteed to find an appropriate dimension for manifolds during the alignment. Numerical simulation and extensive comparisons demonstrate the effectiveness of the derived theorem and proposed method. How to further reduce dependence of our proposal on temporal target predictions is our future work.
\section{Acknowledgments}
This work is supported by the National Natural Science Foundation of China under Grants 61976229, 61976104, 61906046, 61572536, 11631015 and U1611265.

\bigskip

\fontsize{9.0pt}{10.0pt} \selectfont
\bibliography{AAAI-LuoY.3422}
\bibliographystyle{aaai}
\end{document}